# Investigating Literary Motifs in Ancient and Medieval Novels with Large Language Models

Emelie Hallenberg

Department for Linguistics and Philology, Uppsala University

emelie.hallenberg@lingfil.uu.se

## Abstract
The Greek fictional narratives often termed 'love novels' or 'romances', ranging from the first century CE to the middle of the 15th century, have long been considered as similar in many ways, not least in the use of particular literary motifs. By applying the use of fine-tuned large language models, this study aims to investigate which motifs exactly that the texts in this corpus have in common, and in which ways they differ from each other. The results show that while some motifs persist throughout the corpus, others fluctuate in frequency, indicating certain trends or external influences. Conclusively, the method proves to adequately extract literary motifs according to a set definition, providing data for both quantitative and qualitative analyses.

## Introduction

The earliest examples of a certain kind of narrative fiction in Greek prose, later termed 'love novels' and considered important if not primary precursors to the modern novel, date back to the first century CE. The five extant texts were most probably part of a larger, more diverse body of fictional works, of which only fragments remain today (Stephens and Winkler 1996). The latest of the surviving ancient love novels was supposedly written in the fourth century, marking the end of this first period of popularity of the genre, and scholars have researched and debated its origin, supposed audience, contents and style (Whitmarsh 2001; Whitmarsh 2008; Schmeling 1996; Hägg 1983; Anderson 1984; Perry 1967; Reardon 1991; Cueva and Byrne 2014; Karla 2009). In close emulation of these ancient texts, four love novels were composed during the mid-twelfth century in Constantinople, the capital of the Byzantine empire. Although resembling their predecessors in terms of plot, contents and characterization, the Byzantine authors introduced formal elaborations of various kinds – three of them wrote in verse rather than prose – and with varying degrees of subtlety they incorporated elements from Byzantine court milieu and society. A further development in the Greek narrative tradition occurred a couple of centuries later when a number of romances in vernacular rather than Atticizing Greek were written. Influenced by oral storytelling as well as Western chivalric romances, these texts nevertheless have strong connections to the ancient novel tradition. Yet, a systematic and comprehensive exploration of similarities and differences in form and content between these groups of narratives texts remains a desideratum.

In "Forms of time and chronotope in the novel", Russian literary theorist Mikhail Bakhtin wrote that "The plots of these romances (like those of their nearest and most immediate successors, the Byzantine novels) are remarkably similar to each other, and are in fact composed of the very same elements (motifs): individual novels differ from each other only in the number of such elements, their proportionate weight within the whole plot and the way they are combined" (Bakhtin 1981, p. 87). If Bakhtin is correct, these motifs should be easily recognizable, and the ways in which individual novels are different and similar should be possible to spell out and quantify. Therefore, this study aims to investigate *which motifs link ancient and medieval Greek novels together, and which set them apart from each other?* It also asks two main empirical questions:

- Which motifs are most central and predominant in the corpus of ancient and medieval Greek novels? Do these patterns fluctuate or persist over time?
- How do shared motifs link texts together: which texts are more similar, and which are less so? Which motifs are *not* shared between texts?

As we will see, in order to address these questions with a computational approach, some measure of methodological development will be needed to extract and analyze literary motifs in ancient and medieval Greek texts effectively. Accordingly, the results of this study will hopefully also pave the way for related studies in the field of Greek literary genres.

Section 1 outlines a theoretical framework for the study of literary motifs; Section 2 offers a closer description of material and its historical context; Section 3 describes the method; Section 4 the results, followed by an assessment of the investigation and conclusion in Section 5.

## 1. Theoretical background and definitions

Along with the term "theme", "motif" is often used in different ways to denote certain units of meaning in text (along with other synonyms such as "topic", "topos" et cetera). Among literary theorists, diverging views exist on how exactly to define these units – Russian formalists, for example Vladimir Propp and Boris Tomashevsky, regarded motifs (or 'functions' in Propp's terminology) as small textual segments which together constitutes a story (Propp 1968; Tomashevsky 1965). In structuralism, as well, agreement with Tomashevky's definition of motif is prevailing, as it views motifs as smaller constituents that are combined in different ways into larger structures and patterns (Free 1974). Another definition describes motifs as "clusters or families of related words or phrases that, by virtue of their frequency and particular use, tell us something about the author's intentions" (Freedman 1971, p. 123). Theodor Wolpers argues that motifs are all too often neglected and reduced by narratologists to merely actions, and not regarded in their own right (1995).

In the context of the present study, motifs will be defined as units of meaning, extracted from a limited textual segment, that together form part of the narrative structure.

## 2. Material and historical context

The first corpus examined consists of five complete texts, written between the mid-first and fourth centuries CE in Roman Greece and Asia Minor. The texts are *Callirhoe* by Chariton, *Ephesiaca* by Xenophon of Ephesus, *Leucippe and Clitophon* by Achilles Tatius, *Daphnis and Chloe* by Longus, and *Aethiopica* by Heliodorus. The authors are known only by these works, and not much else is known of their lives.

The origin of the novel as a literary form, and how and why it was 'invented', has been the subject of ongoing scholarly debate, with contributions from scholars such as Tomas Hägg (2006), Stefan Tilg (2010), Niklas Holzberg (1994), and Ben E. Perry (1967). In short, these early examples likely arose from a synthesis of contemporary writings – including historiography, drama, travel narratives, poetry and biography – as well as influences from other cultural spheres and oral storytelling traditions.

For approximately eight hundred years thereafter, it has been argued that the role of storytelling in the Greek language was partly filled by hagiography and other religious narratives (Hägg 1983; Elliott 1987). Not until the mid-twelfth century, under the philhellenic Komnenian dynasty, four novels were composed in emulation of the ancient Greek ones within a brief span of a couple of decades. Three of them have survived in complete form: *Rhodanthe and Dosicles* by Theodore Prodromos, *Drosilla and Charicles* by Nicetas Eugenianos and *Hysmine and Hysminias* by Eustathios Macrembolites. The fourth, *Aristhandros and Kallithea* by Constantine Manasses, survives only as fragments.

The authors of these Middle Byzantine novels were elite men within Constantinopolitan court circles, immersed in a cultural environment that read and valued classical literature. This 'renaissance' of the love novel occurred alongside the revival of other ancient genres, such as satire, epic, and historiography (Beaton 1996). The Komnenian novels were long perceived as mere imitations of their Greek predecessors, but have recently been reevaluated and attracted more scholarly interest (Messis, Mullett and Nilsson 2018; Nilsson 2016). While the Byzantine authors relied heavily on their ancient predecessors – using similar themes, motifs, character names, and intertextual elements – they simultaneously molded the stories into forms suitable for their contemporary audiences and socio-cultural context (Roilos 2005). It has been suggested that the Byzantine authors appropriated the novelistic medium, at times surprising or even provoking their audience, to voice contemporary observations and commentaries (MacAlister 1994), as well as incorporating Christian elements (Burton 1998).

The Byzantine romance tradition evolved further under the Palaiologan dynasty between the thirteenth and fifteenth centuries. This collection comprises five romances originally written in Greek: *Livistros and Rhodamni*, *The Tale of Achilles*, *Kallimachos and Chrysorrhoe*, *Velthandros and Chrysantza* and *The Tale of Troy*. Seven additional texts from the same period are adaptations from other sources (six from Western Europe, one from Persia): *Apollonios of Tyre*, *Florios and Platziaflore*, *Old Knight*, *Imperios and Margarona*, *War of Troy*, *Teseida* and *Alexander and Semiramis*. All Palaiologan romances were written anonymously, except for *Kallimachos and Chrysorrhoe*, which was probably written by Andronikos Komnenos Palaiologos, a member of the royal family.

Kostas Yiavis (2019) has questioned this clear division between originals and adaptations in this latter group, suggesting that it may obscure significant overlaps between and differences within the texts. He also emphasizes the diversity of the romances and their sources: "one may identify a historical chronicle, a legend, a distant descendent of a Hellenistic novel, a retake on a chivalric epic, a couple of composite romances[...]" (p. 128). Other scholars have argued that the distinction between originals and adaptations reveals how the Byzantine 'translators' regarded themselves and their audience in relation to the Western society, and how they adapted the source texts so that they would fit into the Byzantine literary system (Vassilopoulou 2024).

Despite their diverse influences, it is certain that at least some of the Palaiologan novels form a continuum within the Greek novelistic tradition, judging by thematic and stylistic similarities (Agapitos 1991). Another relevant issue concerns the paratextual evidence, including title patterns and book divisions – elements that have been convincingly put forward as generic markers by Panagiotis Agapitos (2004). Title conventions have also been discussed by Tim Whitmarsh (2005) who considers them a clear sign of genre awareness among ancient Greek authors. The adoption of the common pattern – the title consisting of the names of the protagonists – by the Komnenian and Palaiologan writers reasonably testifies to their inclusion in the novelistic corpus. For this reason, this study includes only the late Byzantine romances that adhere to this convention.

Thus, in contrast to the Komnenian novels, which were composed during a comparatively short time span and within a restricted geographical area, and to the ancient novels – written over a couple of centuries but targeted at a small, well-educated readership (Bowie 1996) – the vernacular romances stand out in that they were created over approximately two hundred years, and drew material from many different sources. Carolina Cupane (1995) outlines the contemporary spirit as follows: "the love romance of the Palaiologan era, a product of this contradictory period, is characterized equally by a revival of studies of classical texts, which were revisited with true humanistic alacrity and passion, and a cautious but undeniable openness towards the medieval cultural koiné, and it expressively embodies the various currents and opposing trends." Cupane further suggests that the common motifs shared by the late Byzantine romances and Western romances, such as those from Old French literature, imply a direct contact between the two literary systems (Cupane 2019).

Accordingly, the premodern Greek novelistic corpus represents a living tradition with many shared features, unique characteristics, and influences from other cultures. This apparent duality gives rise to numerous open questions and challenges, especially for quantitative approaches aiming to measure these commonalities and differences over time and place.

Research on motifs in these texts has been extensive. An early scholar on the subject, Russian philologist and literary theorist Alexander Veselovsky (1838-1906), uses the word мотивы (motifs) in his seminal work *From the history of the novel and short story* (1886). Veselovsky was a prominent researcher on the circulation and hybridization of themes, motifs, and plots in folktales, myths, and legends, as well as in ancient novels, Christian apocrypha, and medieval epics. One of his theories, particularly relevant to the present study, concerns the reworking and reuse of old motifs in different generic contexts. Veselovsky argues that these motifs reappear when social conditions and circumstances create a need for them. He additionally claims that "Only on the basis of Christian legend did the Greek novel find a continuation" – an intermediary step which he considered crucial in the history of the novel. Regarding the Byzantine novels, he is more critical, for example when it comes to the lack of development of the genre: "The Byzantine romance was an external, academic imitation of the Greek: the same positions and types, the same ideals that life had long left behind"; and the simplicity of style and plot: "the forgetting of style, as well as a tendency toward the simplest fairy-tale plots, for example, in the episodes of Livistros and Rhodamni, and in Callimachos and Chrysorrhoe."

In conclusion, traditional scholarship has not always regarded the Greek novels, and their Byzantine antecedents, favourably in terms of originality, but regardless of position in that debate, the continuum of themes and motifs in these texts have always been acknowledged and stressed.

## 3. Method

How, then, can we identify motifs — defined as recurring units of meaning that appear within limited stretches of text — in a collection of narratives? Besides traditional close reading and recording our interpretations, we may also engage in 'distant reading' practices (see Moretti 2000), particularly by employing various computational methods of natural language processing and visualization to extract and analyze data from a corpus. The study of recurrent or predominant meanings expressed in a body of texts typically involves the use of different varieties of topic modeling, ranging from more basic methods based on word co-occurrences (e.g., Latent Dirichlet Allocation, LDA; see Blei, Ng and Jordan 2003) to more advanced techniques relying on pre-trained sentence embedding models for semantic clustering (e.g., BERTopic; see Grootendorst

2022). In both cases, the outputs are arrays of words and representative topics that require interpretation. The size of the corpus and, in the case of BERTopic, the availability of pre-trained sentence embedding models in the target language are important factors affecting the method's effectiveness. Moreover, the method relies heavily on defining suitable parameters, such as pre-determining the desired number of topics.

Nonetheless, several studies have applied topic modeling techniques to various corpora of literary texts (Navarro-Colorado 2018 for Spanish poetry; Berglund, Dahllöf, and Määttää 2019 for Swedish 21st century prose fiction; Ginn and Hulden 2024 for Roman literature; Martinelli et al 2024 on classic Latin literature; Schöch 2021 on French enlightenment drama). While the method is constantly improving, the fact that it is based on word frequencies and requires some degree of manipulation of the texts in order to get satisfying results (Uglanova and Gius 2020), warrants experimentation with alternative approaches.

The recent rapid development of causal language models trained on vast amounts of data (LLMs) offers new opportunities in computational literary studies. These deep learning models, trained to predict the next token given a series of preceding tokens, can be used for various tasks, including a form of topic modeling, that results not in ambiguous bags of words but cleared natural language labels and free-form descriptions (see e.g. Pham, Sun, Resnik, and Iyyer 2023). However, their effectiveness in dealing specifically with literary motifs and in texts written in various low-resource languages, with only partial representation in their pre-training data — such as ancient Greek – remains an open research question. This study applied the following approach.

The texts were structured in a uniform manner, divided into sentences at full-stops and then chunked into blocks of maximum 1000 tokens according to the OpenAI:s TikTokentokenizer (https://github.com/openai/tiktoken). GPT-4o was fine-tuned on 74 manually annotated examples using the OpenAI:s API (3 batches, batch size 1, Leaning Rate Multiplier 2; result accessible as ft:gpt-4o-2024-08-06:personal:themebot5:AMG1vOu7). Each chunk was paired with the two previous chunks for context (unless it was the first chunk of the novel). The following system prompt was used:

Identify potential literary motifs (recurring recognizable and meaningful patterns of meaning) from the provided text, expressed as concise, single sentences. Focus on motifs related to characters, objects, emotions, or events. Only extract motifs from the current text, ignoring the preceding context. Do not mention character names, and refrain from providing any additional commentary beyond the list of motifs.

The trained model was then used to extract motifs from the whole corpus (as numbered lists of sentences).

The sentences representing the motifs were subsequently embedded using a sentence-transformer (all-mpnet-base-v2), and clustered with BERTopic, UMAP and HDBSCAN.[1]

In the next step, each list of clustered motif labels were summarized by another LLM (Meta's Llama-3.1-8B-Instruct). Some labels tend to become rather complex, since the model tries to capture every aspect of the cluster, which is something to bear in mind when examining the result.

## 4. Results

The evaluation of the model output must be done manually and requires some knowledge about the corpus at hand. An early observation of the extracted motifs, before the rest of the process begins, is crucial in order to end up with satisfying results. These are three examples – one from each sub-corpus – of output from the model.

Source text: *Phlorius and Platzia Phlora*    Φθάνουν τὴν Ἀλεξάνδρειαν, ὀλίγον ἀνασαίνουν. Καὶ πάλιν ἀπεσώσασιν εἰς χώρας Βαβυλῶνος, ὅπου ἦσαν τὰ παλάτια Δαβὶδ τοῦ βασιλέως καὶ ἀπεκεῖ εἰς ξενοδοχειόν, ἐκεῖ ἐξενοδοχίσθην. Τὸν ξενοδόχον ἐρωτᾷ νὰ μάθῃ διὰ τὴν κόρην. Λέγει του: "Ἐξενοδόχησες ἐδῶ ἔμορφον κοράσιον νὰ ἔναι εἰς κάλλος ἐξαίρετος, καλὴ εἰς τὴν θεωρίαν, ὁμάδι νὰ ἔναι ἄρχοντες, ξένοι πραγματευτάδες;". Λέγει του: "Ἐξενοδόχησα, πλὴν τοὺς πραγματευτάδες, ἀλλὰ τὴν κόρην ἤξευρε ὅτι ἐπούλησάν την καὶ ὅσον λογάριν ἔδωκαν, ὅλον ἐτρίπλασάν το· τόσο ἤρεσεν τὸν ἀμιρᾶν τὸ κάλλος τῆς ὡραίας καὶ εἰς ἕνα πύργον ὑψηλὸν ἔβαλεν τὸ κοράσιον. Ὁ πύργος ἔναι δυνατός, γύρωθεν ἔχει κάστρον, ἔχει καὶ καστροφύλακα μὲ προσοχὴν νὰ βλέπῃ, ἔχει καὶ βάγιες ἑκατὸν τὴν κόρην νὰ προσέχουν, ὅλες νὰ βλέπουσιν αὐτήν, τὸ θέλει ἐκείνη νὰ 'ναι".

Translation: I came to Alexandria, and rested a little while. And once again I was saved to the land of Babylon, where the palaces of king David were, and from there to an inn, where I stayed. I asked the innkeeper to learn about the girl. I said: "Did you host a beautiful girl here who was outstanding in beauty, fair in appearance, perhaps with some lords, foreign merchants?" The innkeeper said: "Yes, except that I found out that the merchants sold the girl and however much they gave, she was worth triple. The girl's beauty pleased the Emir so much that he threw her into a high tower. The tower is strong, surrounded by a fortress, and a fortress guard who watches it. It has a hundred guards to watch the girl, and they all look at her, as she wants to be."[2]

Motifs: Arrival in a new city.    Selling of a maiden.    Beauty captivates a ruler.    Maidens guarded in a tower.

---

[1] Hyperparameters for the clustering: UMAP(n_neighbors=5, n_components=5, metric='cosine', min_dist=0.09), HDBSCAN(min_cluster_size=10, min_samples=10, metric='euclidean', cluster_selection_method='eom', prediction_data=True, allow_single_cluster=False

[2] No English translation of this novel exists yet; therefore the translation is mine.

Source text: *Callirhoe*  καὶ προετίμων αὐτὸν πατρίδων καὶ τέκνων. ὁ δὲ μόνας εἴκοσι τριήρεις ἐπελέξατο τὰς ἀρίστας καὶ μεγίστας, ὡς ἂν ὑπὲρ τὸν Ἰόνιον μέλλων περαιοῦσθαι, καὶ ταύταις ἐνεβίβασεν Ἕλληνας μὲν ἅπαντας ὅσοι παρῆσαν, Αἰγυπτίων δὲ καὶ Φοινίκων ὅσους ἔμαθεν εὐζώνους· πολλοὶ καὶ Κυπρίων ἐθελονταὶ ἐνέβησαν. τοὺς δὲ ἄλλους πάντας ἔπεμψεν οἴκαδε, διανείμας αὐτοῖς μέρη τῶν λαφύρων, ἵνα χαίροντες ἐπανίωσι πρὸς τοὺς ἑαυτῶν, ἐντιμότεροι γενόμενοι· καὶ οὐδεὶς ἠτύχησεν οὐδενός, αἰτήσας παρὰ Χαιρέου. Καλλιρόη δὲ προσήνεγκε τὸν κόσμον ἅπαντα τὸν βασιλικὸν Στατείρᾳ.

Translation: They thought more of him than their country and their children. But he chose only twenty ships, the biggest and best, because he was going to cross the Ionian sea; on board these he put all the Greeks who were there and all the Egyptians and Phoenicians he found to be most energetic; many Cypriots embarked too as volunteers. All the rest he sent home; he gave them a share of the spoils so that they could look forward to returning to their familias, because their standing would be improved. No one who had asked anything of Chaereas failed to get it. Callirhoe brought all the royal jewels to Statira. (English translation from Reardon 1989)

Motifs: Preparation for departure.      Recruitment of warriors.      Distribution of spoils.      Recognition of contributions.

Source text: *Hysmine and Hysminias*  Ταῦτ' ἰδὼν τὸν Ἀλκινόου κῆπον ἐδόκουν ὁρᾶν, καὶ μῦθον οὐκ εἶχον τὸ τοῖς ποιηταῖς σεμνολογούμενον πεδίον Ἠλύσιον· δάφνη γὰρ καὶ μυρρίνη καὶ κυπάριττος καὶ ἄμπελοι καὶ τἆλλα τῶν φυτῶν, ὅσα [τὸν κῆπον ἐκόσμει ἢ ἄμπελοι καὶ τἆλλα τῶν φυτῶν, ὅσα [τὸν κῆπον ἐκόσμει ἢ μᾶλλον] ὁ Σωσθένους ἔφερε κῆπος, ἐφαπλοῦσι τοὺς κλάδους ὡς χεῖρας καὶ ὥσπερ χορὸν συστησάμενα κατοροφοῦσι τὸν κῆπον, ἐς τοσοῦτον δὲ τῷ ἡλίῳ παραχωροῦσι προκύψαι περὶ τὴν γῆν, ἐς ὅσον ὁ ζέφυρος πνεύσας τὰ φύλλα διέσεισεν.

Translation: Seeing this, I thought I beheld Alkinoos' garden and felt that I could not take as fiction the Elysian plain so solemnly described by the poets. For laurel and myrtle and cypresses and vines and all the other plants that adorn a garden, or rather that Sosthenes' garden contained, had their branches raised like arms and, as if setting up a dance, they spread a roof over the garden, but they permit the sun to filter through to the ground in as much as the zephyr blew and rustled the leaves. (English translation from Jeffreys 2014)

Motifs: A garden full of beauty and pleasure.    Contest between flowers for beauty.     Plants form a chorus in the garden

Judging by the above and other samples, I determine that the model functions well in its task to extract motifs from text chunks.

The material thus consists of the following data: the Imperial corpus initially contains 8.342 text chunks, the Komnenian 2.803 text chunks, and the Palaiologan 2.891 text chunks. After the exclusion of the chunks that were not clustered to a specific motif (so-called outliers), the numbers for each period were 6.105, 2.100 and 2.214 respectively. This difference in size makes it complicated to look at the absolute number of frequencies for the motifs, which is why I have tried to avoid that.

The size of a motif does not automatically mean that it is significant for the story as such, because it only tells us how many text chunks that contain this specific motif – however, it is still an indication of how much text that is dedicated to the motif, which in itself points to a relative significance. That being said, I wanted to avoid analyses that relied too heavily on numbers and percentages, as these could prove to be misleading. The total number of 350 motifs is to be found in Appendix A.

The questions stated in the introduction will be answered in turn, with regards to the results of these analyses.

*Which motifs are most central and predominant in the corpus? Does this pattern fluctuate or persist over time?*

I constructed two time graphs: one with examples of five fluctuating motifs; and one with more stable motifs. The measurement used for the fluctuating motifs is standard deviation, where a high number indicates a more variance in occurrences between time periods. Two of the most fluctuating motifs actually have relatively high counts in all the corpora, but reach significantly higher numbers in the Komnenian texts: a maiden's beauty and transitions from day to night. An obvious explanation for the first case is the prominence of ekphrastic descriptions – especially of female characters – in the Komnenian novels (see e.g. Holzmeister 2014; Taxidis 2021). The second case could point to the significance of dreams as well as nightly anguishes as a sign of lovesickness in the protagonists (on dreams, see MacAlister 2013).

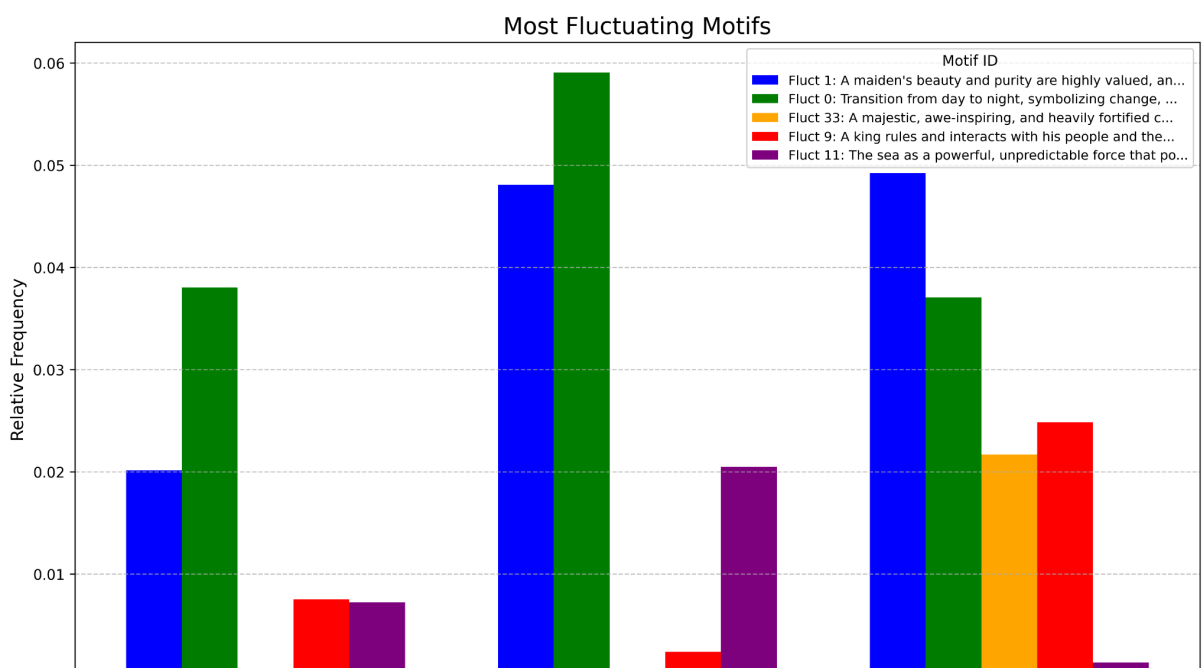

*Figure A. Bar chart visualizing the five most fluctuating motifs between the three time periods. The chart is based on the standard deviation of relative frequency.*

The standard deviation numbers for all of the five most fluctuating motifs are comparatively low (see Appendix D), ranging between 0.0165 and 0.0098, which indicates that few motifs are very prominent in some of the texts while totally absent in others. One exception is the castle, which is a common *topos* in the Palaiologan texts, and is practically absent from the other two time periods (Fonseca 2020). Likewise, the king as a character (possibly a sign of western influence on the Palaiologan romances) is comparatively high in the later texts, where the love god Eros, as well, is portrayed as a king in his castle (on king Eros, see Agapitos 2013; on Western cultural influences on the romances, see Vassilopoulou 2024).

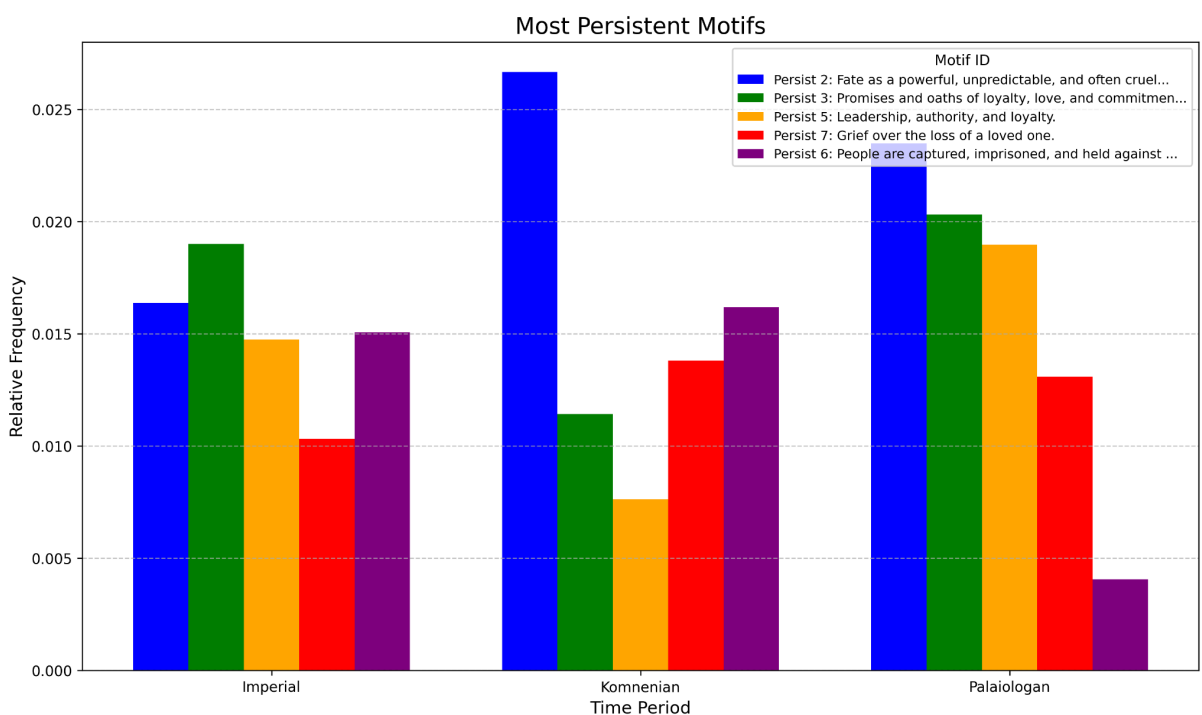

*Figure B. Bar chart visualizing the five most persistent motifs between the three time periods. The chart is based on mean frequency.*

The measurement used for the calculation of the most persistent motifs is mean frequency, that is how much their relative percentage varies between time periods. The most stable of all the motifs, fate as a powerful, unpredictable and often cruel force, has been the focus of much research relating to the novels (see e.g. Van Steen 1998; Chew 2012; Jouanno 2012), but the other four motifs on the top five list have almost equal mean frequency numbers. These motifs are moreover some of the largest, meaning that size and stability seem to go hand in hand, and thereby that none of the motifs are overrepresented in one or two of the texts.

*How do shared motifs link texts together: which are similar and which are less so? Which motifs are not shared between texts?*

In order to measure which novels that are most similar and which that are less similar, their similarity scores were compared. This score is calculated by first representing each novel as a vector (a numerical summary of all the motifs the novel contains); then comparing how closely aligned these vectors are to each other using cosine similarity. The score ranges from 0 to 1, and the higher it is, the more motifs are shared between two texts (see Appendix C for full list of similarity scores).

This analysis shows that the novels that resemble each other the most are *Aithiopica* and *Leucippe and Clitophon* (similarity score 0.81), *Rhodanthe and Dosicles* and *Drosilla and Charicles* (similarity score of 0.80), and *Aithiopica* and *Rhodanthe and Dosicles* (similarity score 0.77). The relationships between some of these texts are interesting: *Rhodanthe and Dosicles* follows the same narrative structure as *Aithiopica* (with a beginning *in medias res* with subsequent flashbacks), as well as similar content. Additionally, *Drosilla and Charicles*, in its turn written in emulation on *Rhodanthe and Dosicles*, is situated close to this very novel (see Jeffreys 2014 on the relationship between the Komnenian novels). The similarity score for these pairs are quite high, indicating an overlap in motifs by around 80%. The upper part of the list with similar novel pairs is dominated by Imperial and Komnenian novels; the first Palaiologan romance on the list is *Livistros and Rhodamni* which shares 64% of its motifs with both *Hysmine and Hysminias* and *Drosilla and Charicles*.

The least similar novels are *Imperios and Margarona* and *Aristandros and Kallithea* (similarity score 0.13) and *Daphnis and Chloe* and *Alexander and Semiramis* (similarity score 0.26). Overall, the lower part of the list is dominated by Palaiologan novels, which not only seem to share less content with the other corpora, but also with each other.

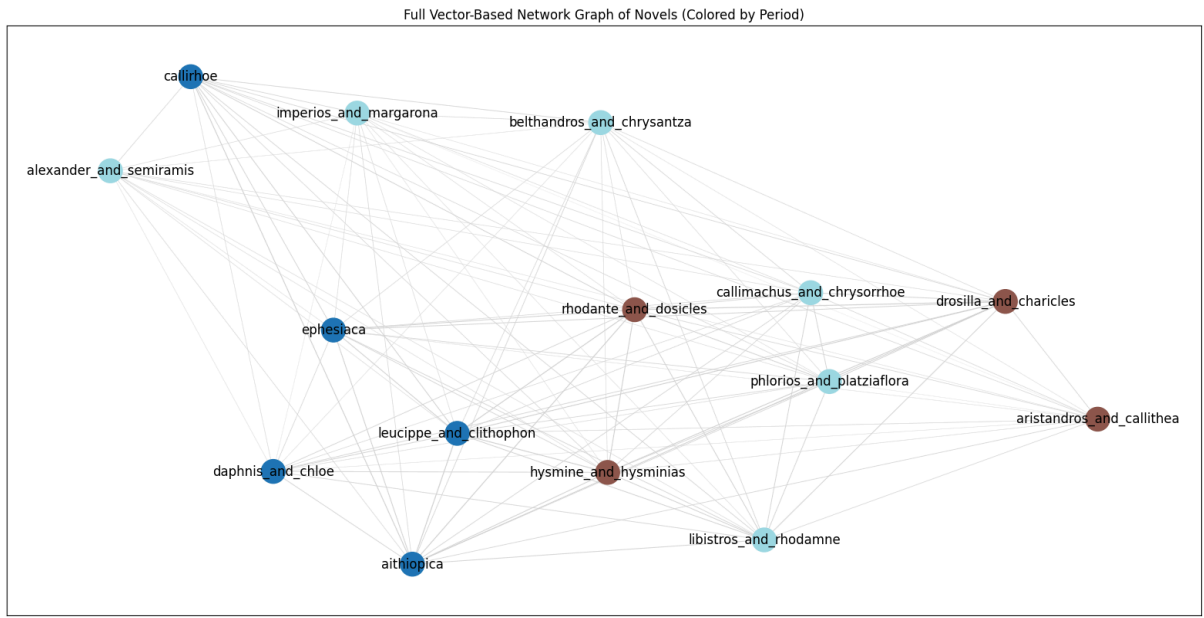

*Figure C. Network-graph visualizing the relationship between the texts, based on vector representations. The distance between two nodes is indicative of how many motifs they share.*

Moving on from how the novels are connected through shared motifs, we will now take a look at the motifs that are distinct for individual novels. The core of the uniqueness score is the ratio between the motif's frequency in a novel and its normalized overall frequency. Because of the variety in size between the three corpora, I made the calculation relative instead of absolute, to prevent any misleading results. The outcome (see Appendix B) shows that several of the novels include motifs that are to some extent unique for them; a fact which highlights the variance that indeed does exist within this corpus, regardless of its, from many aspects, undeniably similar plot structure.

## 5. Conclusion

Here I believe that a few points should be noted considering the application of digital methods to the texts under study. The first vital step comes with the manual annotation of the fine-tuning examples – if these are not properly and adequately done, it will cause undesired results in the final model performance.

Next, the clustering step, which is the part of the process that is most difficult to control – the parameters that are adjustable affect e.g. the size of the clusters, how 'global' or 'local' the structures of the clusters are, and effectively how many clusters that are generated. Nevertheless, examining the largest motifs, regardless of cluster sizes and numbers, they always end up being the same ones, which should indicate that these really are the most common. However, clustering is doubtless the part of the work process where more time could be spent, in attempting to achieve an ideal outcome.

Returning to the definition of motifs previously mentioned, that it should be recurrent, recognizable and contained within a limited amount of text, it is probable that, to get a fair representation of what these motifs really consist of, it would be necessary to examine them thoroughly, not just look at the summarizing label. That, however, is the subject for another study.

The use of language models in this analysis comes with advantages as well as challenges. While the largest benefit undoubtedly is the time gained using distant reading compared to close reading, it naturally entails more unsupervised tasks, performed by the language model. It becomes even more important for researchers to be as careful as possible in declaring their work process and ensure transparency. In terms of replicability, the entire code for this study will be openly available, as is the fine-tuned model, which will enable other scholars to perform the same type of investigation.

In conclusion, I believe this study has shown that a large language model can be helpful in extracting and summarizing important parts of the contents of novels, preparing data for other, more concentrated studies.

Acknowledgements:

The computations and data storage were enabled by resources provided by the National Academic Infrastructure for Supercomputing in Sweden (NAISS), partially funded by the Swedish Research Council through grant agreement no. 2022-06725.

APPENDIX A:

1. Transition from day to night, symbolizing change, uncertainty, and the blurring of reality and dreams. - 438 occurrences

2. A maiden's beauty and purity are highly valued, and she is often the object of desire, love, and protection. - 333 occurrences

3. Fate as a powerful, unpredictable, and often cruel force that determines the course of human life, bringing both suffering and unexpected changes. - 208 occurrences

4. Promises and oaths of loyalty, love, and commitment. - 185 occurrences

5. Hiding one's true identity or intentions through deception, disguise, or secrecy. - 149 occurrences

6. Leadership, authority, and loyalty. - 148 occurrences

7. People are captured, imprisoned, and held against their will. - 135 occurrences

8. Grief over the loss of a loved one. - 121 occurrences

9. Family ties and lineage. - 115 occurrences

10. A king rules and interacts with his people and the world around him. - 106 occurrences

11. Love as an unstoppable, all-encompassing force. - 91 occurrences

12. The sea as a powerful, unpredictable force that poses both threats and opportunities. - 90 occurrences

13. Yearning for an absent or unattainable love. - 88 occurrences

14. Worship and invocation of various goddesses for love, protection, wisdom, beauty, and justice. - 79 occurrences

15. Marriage as a strategic alliance for wealth, power, and social status. - 77 occurrences

16. Preparation for battle. - 74 occurrences

17. Courage in the face of adversity. - 71 occurrences

18. Adornments with gold, jewels, and rich garments signify wealth, status, and power. - 68 occurrences

19. A city under siege. - 65 occurrences

20. Music is a powerful expression of human emotion and connection to the natural world, used in various contexts to convey love, longing, grief, and joy, and to guide, soothe, and unite people - 61 occurrences

21. Offering and sharing wine in various ceremonial and social contexts. - 60 occurrences

22. Unbreakable bonds of friendship formed through shared experiences and tested in adversity. - 60 occurrences

23. Warriors and heroes use various weapons, including swords, bows, spears, and slings, in battles and rituals, often symbolizing courage, fate, and death. - 60 occurrences

24. Creation and care of a beautiful garden. - 59 occurrences

25. Tears, wailing, and sorrowful pleas fill the air as people mourn and lament their suffering, misfortune, and unattainable desires. - 59 occurrences

26. Writing letters as a means of communication and emotional expression. - 59 occurrences

27. Love causes deep emotional pain. - 59 occurrences

28. Conflict between master and servant, often involving deception, betrayal, or disloyalty. - 58 occurrences

29. Divine intervention. - 57 occurrences

30. Death, mourning, and burial. - 57 occurrences

31. A mother grieves over her child's uncertain fate. - 56 occurrences

32. Preparations for and celebration of a wedding. - 56 occurrences

33. Navigating danger with caution and planning. - 55 occurrences

34. Pleading for help or mercy in times of need. - 53 occurrences

35. A majestic, awe-inspiring, and heavily fortified castle, often guarded by a powerful dragon, serves as a symbol of unattainable desire, beauty, and power. - 51 occurrences

36. Eyes convey emotions, intentions, and connections through direct and intense gazes. - 50 occurrences

37. Love expressed through tender, passionate, and intimate physical contact. - 49 occurrences

38. False accusations and confessions of wrongdoing. - 48 occurrences

39. Being forced into servitude or slavery against one's will. - 48 occurrences

40. Human perseverance in the face of adversity. - 48 occurrences

41. Separation from family and loved ones. - 47 occurrences

42. Separation of lovers due to external forces. - 47 occurrences

43. Youth, beauty, and passion. - 46 occurrences

44. Unwavering commitment and loyalty in the face of adversity, demonstrated through actions such as sacrifice, guardianship, and willingness to risk one's life for the beloved. - 46 occurrences

45. A friend shares in and alleviates another's suffering. - 45 occurrences

46. A woman's commitment to preserving her virginity as a test of her purity, virtue, and chastity in the face of danger, temptation, and societal pressure. - 45 occurrences

47. Wealth and its consequences. - 45 occurrences

48. Farewell to loved ones. - 45 occurrences

49. Maintaining silence in the presence of significant events, emotions, or authority. - 44 occurrences

50. A skilled shepherd, accompanied by loyal dogs, leads and protects his flock with the help of music and his knowledge of shepherding. - 44 occurrences

51. Enduring a long, arduous journey through diverse and often treacherous landscapes, including dense forests and mountains, in search of survival, rescue, or a new life. - 43 occurrences

52. A ring symbolizing love, protection, and connection. - 43 occurrences

53. Encounter with a stranger. - 43 occurrences

54. Women navigating complex relationships and societal expectations. - 43 occurrences

55. Forced or unwanted marriage. - 43 occurrences

56. Rituals performed to honor and connect with deities, ancestors, and the community through ceremonial acts, purification, and sacrifices. - 42 occurrences

57. Parents mourn the loss of their child. - 42 occurrences

58. A character experiences intense emotional pain and sorrow, often due to loss, separation, or betrayal, and seeks comfort, solace, or pity from others. - 42 occurrences

59. Preparation and sharing of a lavish meal. - 42 occurrences

60. Connection between nature and human emotions. - 41 occurrences

61. Divine guidance through prophecy. - 41 occurrences

62. Fear of divine retribution. - 40 occurrences

63. Regalia and symbols of divine or royal authority. - 40 occurrences

64. Forced removal from one's homeland. - 40 occurrences

65. Punishment and death by various means, including fire, crucifixion, beheading, and stoning. - 39 occurrences

66. A woman of exceptional beauty, power, and authority faces challenges, makes difficult decisions, and evokes strong emotions in those around her. - 39 occurrences

67. Pursuit and capture by pirates. - 38 occurrences

68. Humans engage in various forms of hunting and fishing for sustenance, skill, and enjoyment, often with the aid of dogs, horses, and specialized tools. - 38 occurrences

69. Leaders interact with others for guidance, support, and decision-making. - 38 occurrences

70. Jealousy drives individuals to extreme actions, often leading to conflict, betrayal, and destruction. - 37 occurrences

71. Departure in secrecy, often on horseback, between the city and the countryside. - 37 occurrences

72. Victory celebration. - 36 occurrences

73. Struggle to maintain virtue and control in the face of conflicting desires and emotions. - 36 occurrences

74. A woman mourns her loss of freedom and happiness. - 36 occurrences

75. Calling on higher powers. - 36 occurrences

76. Righteous judgment by a discerning, incorruptible, and unforgiving authority. - 36 occurrences

77. Unrequited love. - 35 occurrences

78. Roses and flowers symbolize love, beauty, and nature. - 35 occurrences

79. A person flees or departs from a place. - 35 occurrences

80. Clashes between opposing forces. - 35 occurrences

81. River as a source of life, nourishment, and sustenance. - 34 occurrences

82. Romantic pursuit. - 34 occurrences

83. Building grand, lavish tombs near the sea. - 34 occurrences

84. Fear in the face of unknown or perceived threats. - 33 occurrences

85. Association with Greek gods and mythic heroes. - 33 occurrences

86. Hands raised to heaven in supplication, offering libations and prayers to various gods and goddesses for salvation, protection, blessings, and divine favor. - 33 occurrences

87. Interactions between rulers and their subjects. - 33 occurrences

88. Sacrificing one's own interests, freedom, or possessions for the benefit of a loved one or a higher cause. - 32 occurrences

89. War and foreign domination threaten the homeland, bringing unexpected events, various forms of warfare, and the risk of war, but also opportunities for greatness through peaceful negotiations and the mingling of war and peace. - 32 occurrences

90. Precious stones of value and power. - 32 occurrences

91. Birds. - 31 occurrences

92. Love leads to entanglement and captivity. - 31 occurrences

93. A person is torn between their love for someone and the consequences that come with it, leading to inner conflict and turmoil. - 31 occurrences

94. Performing sacrificial rituals to appease gods, honor the dead, and mark significant events. - 31 occurrences

95. Journeys across the sea, arrivals, departures, and encounters with ships. - 31 occurrences

96. A crowd gathers. - 31 occurrences

97. Bandits terrorize and plunder communities. - 30 occurrences

98. A skilled craftsman creates beautiful, intricate, and meaningful works of art that reveal truth and express devotion, love, and emotions through the finest elements of the world. - 30 occurrences

99. Hero embarks on a perilous journey to rescue and save others. - 30 occurrences

100. Presenting offerings to divine figures. - 30 occurrences

101. Seeking help from supernatural or divine beings for healing or guidance. - 30 occurrences

102. Tears and sighs symbolize emotional struggle and sorrow. - 29 occurrences

103. Ritual sacrifice at a sacred or unholy altar. - 29 occurrences

104. Loss of consciousness due to overwhelming emotional stress. - 29 occurrences

105. Exposure and loss of dignity. - 29 occurrences

106. Betrayal of love and trust. - 28 occurrences

107. Eunuchs as loyal, trusted servants and messengers who hold significant influence and often serve as guardians, advisors, and intermediaries in secretive matters under the ruler's command. - 28 occurrences

108. Raiders destroy homes, crops, livestock, and people, leaving desolation and death in their wake. - 28 occurrences

109. Father-daughter relationship. - 28 occurrences

110. Fire symbolizes intense, consuming passion that can lead to destruction, loss, and divine retribution, but also represents a bond, love, and control over nature. - 28 occurrences

111. Breaking free from restraints. - 28 occurrences

112. Seeking comfort and relief from hardship and danger. - 28 occurrences

113. Protection and pursuit of a woman. - 28 occurrences

114. Guidance and recognition through light. - 28 occurrences

115. Inner turmoil. - 27 occurrences

116. Leaders are honored and praised by their followers. - 27 occurrences

117. Entrance and exit through guarded gates. - 27 occurrences

118. A father arranges his daughter's marriage. - 27 occurrences

119. Betrayal. - 27 occurrences

120. Admiration of exceptional beauty. - 27 occurrences

121. Abduction of a maiden. - 27 occurrences

122. Joy and sorrow intertwined. - 26 occurrences

123. Despair in a desolate, sorrowful world. - 26 occurrences

124. A crowd reacts with a mixture of emotions amidst noise and commotion. - 26 occurrences

125. Metamorphosis into a bird or animal form. - 26 occurrences

126. Searching for a lost loved one. - 26 occurrences

127. Punishment meted out by those in power. - 26 occurrences

128. Divine retribution. - 25 occurrences

129. Bringing the dead back to life. - 25 occurrences

130. Painful separation from loved ones. - 25 occurrences

131. Divine beings interact with mortals in natural settings. - 25 occurrences

132. Mourning a lost loved one. - 25 occurrences

133. Gifts are exchanged as symbols of honor, gratitude, love, and connection. - 25 occurrences

134. Wealthy, high-ranking noble or ruler. - 25 occurrences

135. Resistance to oppressive authority. - 25 occurrences

136. Contemplation of self-destruction. - 25 occurrences

137. Honoring and living up to noble lineage. - 25 occurrences

138. Laurel-adorned crowns and wreaths symbolize honor, purity, and virginity. - 25 occurrences

139. Divine intervention in human life. - 24 occurrences

140. Fear of love's uncontrollable power. - 24 occurrences

141. Sharing personal stories with others. - 24 occurrences

142. Betrayal by those closest to you. - 24 occurrences

143. Beauty lost or fading due to inner turmoil or sorrow. - 24 occurrences

144. Conflict between duty and personal desire. - 24 occurrences

145. Despair in the face of overwhelming adversity. - 23 occurrences

146. Confrontation with a wild or ferocious beast. - 23 occurrences

147. Beauty captivates. - 23 occurrences

148. Traveling by sea. - 23 occurrences

149. Authority, loyalty, and power dynamics between a mistress and those serving or vying for her attention. - 23 occurrences

150. A young man of exceptional strength and beauty faces challenges and adversity. - 23 occurrences

151. Expressing gratitude through praise and offerings to gods and goddesses. - 22 occurrences

152. A powerful bird of prey seizes or takes something of value. - 22 occurrences

153. Warrior in combat. - 22 occurrences

154. Sudden intense emotional overwhelm. - 22 occurrences

155. Finding or creating a safe and peaceful place to escape to. - 22 occurrences

156. A letter is written, sent, and received, causing emotional turmoil and internal conflict about its truth. - 22 occurrences

157. Making amends and seeking forgiveness. - 22 occurrences

158. Healing through love and nature. - 22 occurrences

159. Struggle for power within a leader. - 21 occurrences

160. Mighty, hybrid creature with animal attributes, rooted feet, and a long, thick tail, embodying strength, resilience, and longevity, often associated with divine or magical powers. - 21 occurrences

161. Preparation and display of luxurious items for special occasions. - 21 occurrences

162. Reverence for divine authority. - 21 occurrences

163. Beauty compared to nature. - 21 occurrences

164. Divine arrangement of marriage. - 21 occurrences

165. Defiance against oppressive power. - 21 occurrences

166. Relaxation and rejuvenation through immersion in natural water sources. - 20 occurrences

167. Despair over loss of a loved one. - 20 occurrences

168. Sacrifice to appease the sea god. - 20 occurrences

169. Caring for and being cared for by animals, particularly goats. - 20 occurrences

170. Devotion to a loved one or deity. - 20 occurrences

171. Suitors gather, a marriage is proposed, and preparations are made. - 20 occurrences

172. Voyage preparation and defense. - 20 occurrences

173. Journey to a sacred temple for refuge and ritual. - 20 occurrences

174. Sacrificial feast and communal celebration. - 19 occurrences

175. Merchants engage in trade and commerce. - 19 occurrences

176. Love and beauty intertwined with nature and symbolized by animals, trees, and natural elements. - 19 occurrences

177. Letters are exchanged between characters. - 19 occurrences

178. A sleeping maiden surrounded by silence and stillness. - 19 occurrences

179. A hero confronts and defeats a powerful, devouring dragon that poses a threat to the kingdom's survival. - 19 occurrences

180. Reunion with loved ones after separation. - 19 occurrences

181. Competing to show respect and kindness to guests. - 19 occurrences

182. Love and anger are inextricably linked, often residing in close proximity, and can transform into each other in response to rejection, scorn, or unrequited feelings. - 19 occurrences

183. Preparation and administration of poisonous substances. - 19 occurrences

184. Love as an all-consuming flame. - 19 occurrences

185. Finding peace and happiness after overcoming challenges. - 19 occurrences

186. Demand for fair judgment and punishment for those who commit sacrilege and murder. - 19 occurrences

187. Influencing others through various means to achieve a desired outcome. - 19 occurrences

188. Faith in divine guidance and protection. - 19 occurrences

189. Fragrant aromas fill the air. - 19 occurrences

190. Investigation and judgment by a priest. - 19 occurrences

191. Divine favor is granted to those who receive the gods' assistance, approval, and benevolence. - 19 occurrences

192. Love and duty in conflict. - 19 occurrences

193. Maintaining Greek identity in foreign lands. - 19 occurrences

194. Begging for divine assistance. - 18 occurrences

195. Divine beauty that overwhelms and inspires awe. - 18 occurrences

196. Tearing or cutting of one's hair, clothing, or body parts as a physical expression of intense emotional pain, loss, or transformation. - 18 occurrences

197. A powerful, magical, and often treacherous old woman uses her cunning, sorcery, and dark magic to manipulate and deceive others. - 18 occurrences

198. Abundance and provision of food, shelter, and wealth. - 18 occurrences

199. A youthful figure, often a child, is revered for their innocence and purity, believed to possess divine qualities or be of divine origin, and serves as a guide or inspiration for others. - 18 occurrences

200. Embracing and kissing someone. - 18 occurrences

201. Fear of violent or untimely death. - 18 occurrences

202. A woman's extraordinary beauty captivates and silences a crowd. - 18 occurrences

203. Divine forces shape life outcomes. - 18 occurrences

204. Secret sale of a woman. - 18 occurrences

205. Voyage through treacherous weather. - 18 occurrences

206. Encountering a familiar figure that evokes strong emotions. - 18 occurrences

207. Servants providing care and assistance. - 18 occurrences

208. Invoking and honoring Aphrodite through various rituals and prayers for love, protection, and sustenance. - 17 occurrences

209. Sacrificing personal items or body parts to fulfill a demand or make a vow. - 17 occurrences

210. Gifts and tokens exchanged between people. - 17 occurrences

211. Crying. - 17 occurrences

212. A grand, opulent feast prepared by servants, overseen by a master or a woman, signifying alliance, love, or marriage. - 17 occurrences

213. The unbearable weight of unending sorrow. - 17 occurrences

214. Acknowledging and reciprocating kindness and contributions. - 17 occurrences

215. Seeking guidance and aid from a respected elder in a time of need. - 17 occurrences

216. Desperate fight for freedom. - 17 occurrences

217. Judgment by a ruler or leader. - 17 occurrences

218. Discovery of a lost or abandoned child. - 17 occurrences

219. Worship at divine altars. - 17 occurrences

220. Love as a double-edged, unpredictable, and potentially destructive force that can bring both harm and healing. - 17 occurrences

221. Beauty exacts a price. - 16 occurrences

222. Access and control through magical objects and incantations. - 16 occurrences

223. Humiliation through mocking words and actions. - 16 occurrences

224. Confronting past mistakes and their consequences. - 16 occurrences

225. Disguise or deception through illness or injury. - 16 occurrences

226. Betrayal of trust by those who are supposed to be loyal. - 16 occurrences

227. Wisdom in adversity. - 16 occurrences

228. Struggling to survive in a desperate attempt to escape danger. - 16 occurrences

229. Assembling and mobilizing forces for battle or defense. - 16 occurrences

230. Concealed romantic relationships. - 16 occurrences

231. Sacrifice to the gods. - 16 occurrences

232. Fear of being punished or abandoned by those in power. - 16 occurrences

233. Pursuit and capture leading to loss of freedom and threat of death. - 16 occurrences

234. Interconnected growth and abundance through cultivation and care. - 16 occurrences

235. Tears of sorrow. - 16 occurrences

236. Seeking and receiving guidance from trusted authorities. - 16 occurrences

237. Tears overflowing like a river. - 16 occurrences

238. Envy of others' success and prosperity. - 16 occurrences

239. Divine intervention in human affairs. - 15 occurrences

240. Deadly projectiles launched by skilled marksmen. - 15 occurrences

241. Internal conflict between desire and restraint. - 15 occurrences

242. False accusation of murder. - 15 occurrences

243. A woman's unbridled anger towards a male figure. - 15 occurrences

244. A parent's blessing for their child's journey. - 15 occurrences

245. Elephants in battle. - 15 occurrences

246. Loss of a loved one due to separation or destruction. - 15 occurrences

247. Final preparations before embarking on a perilous journey. - 15 occurrences

248. A serene natural landscape with lush vegetation and flowing water. - 15 occurrences

249. Fear of punishment and consequences through capture, trial, and execution. - 15 occurrences

250. Leaving one's homeland. - 15 occurrences

251. Holding onto cherished memories of love, family, and beauty in the face of sorrow and loss. - 15 occurrences

252. Calling on gods for help in times of need. - 15 occurrences

253. A lover mourns the loss of a loved one. - 15 occurrences

254. Abduction of a young girl. - 15 occurrences

255. Pursuit of a fugitive through treacherous terrain. - 15 occurrences

256. Return home. - 15 occurrences

257. Ritual gathering around a lit flame. - 15 occurrences

258. Desire for retribution against those who have wronged or disrespected. - 15 occurrences

259. Marriage and death are inextricably linked. - 15 occurrences

260. Journeying across or towards a body of water, often in a state of urgency or transition. - 15 occurrences

261. Longing for someone who does not reciprocate affection. - 14 occurrences

262. Loyalty to family. - 14 occurrences

263. Enduring prolonged physical and emotional torment. - 14 occurrences

264. Acceptance of inescapable fate. - 14 occurrences

265. Forced descent into a body of water. - 14 occurrences

266. Enduring hardships to find solace. - 14 occurrences

267. Appeal to the sea and its gods for power, comfort, and protection. - 14 occurrences

268. Love ignites and transforms the heart. - 14 occurrences

269. Enforcement of laws to maintain order and punish transgressions. - 14 occurrences

270. Fear of being overwhelmed and lost at sea. - 14 occurrences

271. Planning and executing a secret departure. - 14 occurrences

272. Exploration and reverence for the Nile river. - 14 occurrences

273. Youthful beauty is short-lived and admired. - 14 occurrences

274. Walking a tightrope between life and death. - 14 occurrences

275. Unfulfilled or reckless pursuit of desire leads to suffering and destruction. - 14 occurrences

276. Young people seek guidance and aid from a wise old woman who shares her home, food, healing, and prayers for their safety and well-being. - 14 occurrences

277. A father provides guidance, care, and reassurance to his child. - 14 occurrences

278. Sacrifices are demanded and offered to appease the gods. - 14 occurrences

279. Authority of the father. - 13 occurrences

280. Burning flames of intense passion. - 13 occurrences

281. Overwhelming emotional release through tears. - 13 occurrences

282. Gods are invoked for protection, guidance, and assistance in times of need. - 13 occurrences

283. Loss of freedom through overpowering love. - 13 occurrences

284. Fear of losing control and being at the mercy of uncontrollable forces. - 13 occurrences

285. Hiding in secluded natural areas. - 13 occurrences

286. Love and death are inextricably entwined, often resulting in the loss of a loved one. - 13 occurrences

287. Physical beauty is valued as a measure of worth. - 13 occurrences

288. Preserving feminine purity through restraint and modest behavior. - 13 occurrences

289. A beautiful, skillfully crafted, and mysterious water fountain that symbolizes purity, elegance, and beauty, often found in unexpected places, such as the desert, and associated with the goddess of love, - 13 occurrences

290. A crowd shows approval through loud shouts and cheers. - 13 occurrences

291. Intense, overwhelming grief. - 13 occurrences

292. Compassionate care for those in need. - 13 occurrences

293. Surprise nighttime attack on an unsuspecting village or household. - 13 occurrences

294. Armies assemble and prepare for battle. - 13 occurrences

295. Betrayal or treachery is suspected or accused. - 13 occurrences

296. Writing a message on an arrow to convey love. - 13 occurrences

297. A woman's extraordinary beauty captivates and surpasses all others. - 13 occurrences

298. Battle between Egyptian and Persian forces. - 13 occurrences

299. Healing past wounds. - 12 occurrences

300. A young woman makes a grand entrance into a chamber. - 12 occurrences

301. Violent desecration of sacred places. - 12 occurrences

302. Suppressing and hiding true emotions, especially in the presence of others. - 12 occurrences

303. Surrender to a higher power. - 12 occurrences

304. Captives are bound by heavy chains. - 12 occurrences

305. Forbidden love between siblings. - 12 occurrences

306. Intimate connection between two people. - 12 occurrences

307. Reluctance to speak truthfully due to fear. - 12 occurrences

308. Abundant water flowing from a natural source. - 12 occurrences

309. Escape from a besieged, war-torn castle ruled by a mad king. - 12 occurrences

310. Romantic infatuation. - 12 occurrences

311. Protection of women's virtue. - 12 occurrences

312. A beautiful girl is admired for her exceptional beauty and virtue. - 12 occurrences

313. A token is exchanged as a sign of recognition. - 12 occurrences

314. Sudden alarm. - 12 occurrences

315. Laughter. - 12 occurrences

316. Hidden sorrows are revealed through discreet tears that change appearance, transforming the weeper's outer appearance and evoking pity in the viewer's eyes. - 12 occurrences

317. Purification and protection through devotion to Artemis. - 12 occurrences

318. Bittersweet goodbye kiss. - 12 occurrences

319. Preparation for departure. - 12 occurrences

320. A magical golden apple. - 11 occurrences

321. A malevolent force disrupts and destroys everything in its path. - 11 occurrences

322. Youthful human beauty is favored by gods. - 11 occurrences

323. Betrayal of a maiden by a treacherous female figure. - 11 occurrences

324. Desire for death as a means to escape suffering or misfortune. - 11 occurrences

325. Rumors spread quickly. - 11 occurrences

326. Echoing voices in a natural setting. - 11 occurrences

327. Nymphs nurture and protect children in harmony with nature. - 11 occurrences

328. Regal, ornate thrones atop grand structures. - 11 occurrences

329. Digging beneath obstacles to create hidden pathways. - 11 occurrences

330. A search for a mysterious woman. - 11 occurrences

331. A character observes another with a mix of curiosity and amazement. - 11 occurrences

332. Forced or arranged union of a young person, often a girl, to a partner chosen by an authority figure, such as a father or ruler. - 11 occurrences

333. Weeping. - 11 occurrences

334. A small, swift, and skilled Greek force, united in valor and spirit, defeats invaders through military strength and heroism, bringing glory to Greece. - 11 occurrences

335. Plotting against others. - 11 occurrences

336. Regretful contemplation of past choices. - 11 occurrences

337. Nature consoles and mourns with the sorrowful. - 11 occurrences

338. A maiden offers or receives a cup. - 11 occurrences

339. Maintaining social standing through careful navigation of relationships and actions. - 11 occurrences

340. Love conveyed through written declarations. - 11 occurrences

341. Guiding winds for safe and swift travel. - 11 occurrences

342. Crossing a body of water. - 11 occurrences

343. Instant, unbreakable bonds formed through shared experiences and actions. - 11 occurrences

344. Stealthy encounters under the cover of darkness. - 11 occurrences

345. Communal consumption of food and drink. - 11 occurrences

346. Love is a heavy burden. - 10 occurrences

347. A small, isolated landmass surrounded by water. - 10 occurrences

348. A young man falls deeply in love. - 10 occurrences

349. Grief overwhelms the heart. - 10 occurrences

350. Revival through cleansing with water. - 10 occurrences

APPENDIX B: Top 3 most unique motifs for each novel

    Novel: Aithiopica
     - Unique Topic 38: Human perseverance in the face of adversity. (Relative Uniqueness Score: 2.59)
     - Unique Topic 25: Preparation for battle. (Relative Uniqueness Score: 2.46)
     - Unique Topic 17: A city under siege. (Relative Uniqueness Score: 2.09)

    Novel: Leucippe and Clitophon
     - Unique Topic 37: False accusations and confessions of wrongdoing. (Relative Uniqueness Score: 4.10)
     - Unique Topic 13: Worship and invocation of various goddesses for love, protection, wisdom, beauty, and justice. (Relative Uniqueness Score: 2.54)
     - Unique Topic 35: Love expressed through tender, passionate, and intimate physical contact. (Relative Uniqueness Score: 2.53)

    Novel: Ephesiaca
     - Unique Topic 30: Preparations for and celebration of a wedding. (Relative Uniqueness Score: 3.89)
     - Unique Topic 26: Tears, wailing, and sorrowful pleas fill the air as people mourn and lament their suffering, misfortune, and unattainable desires. (Relative Uniqueness Score: 2.85)
     - Unique Topic 13: Worship and invocation of various goddesses for love, protection, wisdom, beauty, and justice. (Relative Uniqueness Score: 2.84)

    Novel: Callirhoe
     - Unique Topic 28: Death, mourning, and burial. (Relative Uniqueness Score: 4.85)
     - Unique Topic 13: Worship and invocation of various goddesses for love, protection, wisdom, beauty, and justice. (Relative Uniqueness Score: 3.16)
     - Unique Topic 37: False accusations and confessions of wrongdoing. (Relative Uniqueness Score: 2.81)

    Novel: Rhodanthe and Dosicles
     - Unique Topic 15: Courage in the face of adversity. (Relative Uniqueness Score: 2.54)
     - Unique Topic 14: Marriage as a strategic alliance for wealth, power, and social status. (Relative Uniqueness Score: 2.35)
     - Unique Topic 11: The sea as a powerful, unpredictable force that poses both threats and opportunities. (Relative Uniqueness Score: 2.05)

    Novel: Callimachus and Chrysorrhoe
     - Unique Topic 33: A majestic, awe-inspiring, and heavily fortified castle, often guarded by a powerful dragon, serves as a symbol of unattainable desire, beauty, and power. (Relative Uniqueness Score: 5.27)
     - Unique Topic 24: Creation and care of a beautiful garden. (Relative Uniqueness Score: 4.53)
     - Unique Topic 25: Preparation for battle. (Relative Uniqueness Score: 2.13)

    Novel: Alexander and Semiramis
     - Unique Topic 5: Leadership, authority, and loyalty. (Relative Uniqueness Score: 3.37)
     - Unique Topic 16: Adornments with gold, jewels, and rich garments signify wealth, status, and power. (Relative Uniqueness Score: 3.05)
     - Unique Topic 15: Courage in the face of adversity. (Relative Uniqueness Score: 2.90)

Novel: Drosilla and Charicles
- Unique Topic 21: Love causes deep emotional pain. (Relative Uniqueness Score: 2.34)
- Unique Topic 6: People are captured, imprisoned, and held against their will. (Relative Uniqueness Score: 2.34)
- Unique Topic 26: Tears, wailing, and sorrowful pleas fill the air as people mourn and lament their suffering, misfortune, and unattainable desires. (Relative Uniqueness Score: 2.11)

Novel: Imperios and Margarona
- Unique Topic 25: Preparation for battle. (Relative Uniqueness Score: 7.16)
- Unique Topic 28: Death, mourning, and burial. (Relative Uniqueness Score: 4.39)
- Unique Topic 16: Adornments with gold, jewels, and rich garments signify wealth, status, and power. (Relative Uniqueness Score: 3.70)

Novel: Daphnis and Chloe
- Unique Topic 18: Music is a powerful expression of human emotion and connection to the natural world, used in various contexts to convey love, longing, grief, and joy, and to guide, soothe, and unite people (Relative Uniqueness Score: 7.46)
- Unique Topic 35: Love expressed through tender, passionate, and intimate physical contact. (Relative Uniqueness Score: 3.04)
- Unique Topic 24: Creation and care of a beautiful garden. (Relative Uniqueness Score: 2.93)

Novel: Velthandros and Chrysantza
- Unique Topic 33: A majestic, awe-inspiring, and heavily fortified castle, often guarded by a powerful dragon, serves as a symbol of unattainable desire, beauty, and power. (Relative Uniqueness Score: 3.71)
- Unique Topic 28: Death, mourning, and burial. (Relative Uniqueness Score: 3.52)
- Unique Topic 34: Eyes convey emotions, intentions, and connections through direct and intense gazes. (Relative Uniqueness Score: 2.76)

Novel: Hysmine and Hysminias
- Unique Topic 36: Being forced into servitude or slavery against one's will. (Relative Uniqueness Score: 3.06)
- Unique Topic 23: Offering and sharing wine in various ceremonial and social contexts. (Relative Uniqueness Score: 2.98)
- Unique Topic 35: Love expressed through tender, passionate, and intimate physical contact. (Relative Uniqueness Score: 2.61)

Novel: Livistros and Rhodamni
- Unique Topic 20: Writing letters as a means of communication and emotional expression. (Relative Uniqueness Score: 4.60)
- Unique Topic 21: Love causes deep emotional pain. (Relative Uniqueness Score: 2.20)
- Unique Topic 33: A majestic, awe-inspiring, and heavily fortified castle, often guarded by a powerful dragon, serves as a symbol of unattainable desire, beauty, and power. (Relative Uniqueness Score: 2.13)

Novel: Phlorius and Platzia Flora
- Unique Topic 9: A king rules and interacts with his people and the world around him. (Relative Uniqueness Score: 3.38)

  - Unique Topic 1: A maiden's beauty and purity are highly valued, and she is often the object of desire, love, and protection. (Relative Uniqueness Score: 2.75)

  - Unique Topic 22: Unbreakable bonds of friendship formed through shared experiences and tested in adversity. (Relative Uniqueness Score: 2.21)

  Novel: Aristandros and Callithea

  - Unique Topic 22: Unbreakable bonds of friendship formed through shared experiences and tested in adversity. (Relative Uniqueness Score: 4.50)

  - Unique Topic 2: Fate as a powerful, unpredictable, and often cruel force that determines the course of human life, bringing both suffering and unexpected changes. (Relative Uniqueness Score: 3.66)

  - Unique Topic 34: Eyes convey emotions, intentions, and connections through direct and intense gazes. (Relative Uniqueness Score: 3.44)

## APPENDIX C:

Novel pairs sorted by similarity scores (most similar at the top):

Aithiopica and Leucippe and Clitophon: similarity 0.81

Rhodanthe and Dosicles and Drosilla and Charicles: similarity 0.80

Aithiopica and Rhodanthe and Dosicles: similarity 0.77

Leucippe and Clitophon and Rhodanthe and Dosicles: similarity 0.77

Leucippe and Clitophon and Hysmine and Hysminias: similarity 0.74

Rhodanthe and Dosicles and Hysmine and Hysminias: similarity 0.74

Aithiopica and Callirhoe: similarity 0.73

Leucippe and Clitophon and Drosilla and Charicles: similarity 0.72

Drosilla and Charicles and Hysmine and Hysminias: similarity 0.72

Aithiopica and Drosilla and Charicles: similarity 0.69

Leucippe and Clitophon and Ephesiaca: similarity 0.69

Aithiopica and Ephesiaca: similarity 0.68

Aithiopica and Hysmine and Hysminias: similarity 0.66

Ephesiaca and Rhodanthe and Dosicles: similarity 0.66

Hysmine and Hysminias and Livistros and Rhodamni: similarity 0.64

Drosilla and Charicles and Livistros and Rhodamni: similarity 0.64

Rhodanthe and Dosicles and Phlorius and Platzia Flora: similarity 0.64

Rhodanthe and Dosicles and Livistros and Rhodamni: similarity 0.63

Aithiopica and Livistros and Rhodamni: similarity 0.63

Leucippe and Clitophon and Livistros and Rhodamni: similarity 0.63

Daphnis and Chloe and Hysmine and Hysminias: similarity 0.63

Ephesiaca and Drosilla and Charicles: similarity 0.62

Leucippe and Clitophon and Callirhoe: similarity 0.62

Callimachus and Chrysorrhoe and Drosilla and Charicles: similarity 0.62

Leucippe and Clitophon and Daphnis and Chloe: similarity 0.61

Drosilla and Charicles and Phlorius and Platzia Flora: similarity 0.61

Callimachus and Chrysorrhoe and Livistros and Rhodamni: similarity 0.59

Callirhoe and Rhodanthe and Dosicles: similarity 0.59

Callimachus and Chrysorrhoe and Phlorius and Platzia Flora: similarity 0.58

Livistros and Rhodamni and Phlorius and Platzia Flora: similarity 0.58

Rhodanthe and Dosicles and Callimachus and Chrysorrhoe: similarity 0.57

Ephesiaca and Callirhoe: similarity 0.56

Aithiopica and Daphnis and Chloe: similarity 0.56

Ephesiaca and Livistros and Rhodamni: similarity 0.55

Daphnis and Chloe and Livistros and Rhodamni: similarity 0.55

Drosilla and Charicles and Daphnis and Chloe: similarity 0.55

Rhodanthe and Dosicles and Daphnis and Chloe: similarity 0.54

Callirhoe and Velthandros and Chrysantza: similarity 0.54

Ephesiaca and Hysmine and Hysminias: similarity 0.54

Aithiopica and Callimachus and Chrysorrhoe: similarity 0.53

Hysmine and Hysminias and Phlorius and Platzia Flora: similarity 0.53

Callirhoe and Livistros and Rhodamni: similarity 0.52

Drosilla and Charicles and Aristandros and Callithea: similarity 0.51

Velthandros and Chrysantza and Livistros and Rhodamni: similarity 0.51

Aithiopica and Velthandros and Chrysantza: similarity 0.51

Aithiopica and Aristandros and Callithea: similarity 0.50

Ephesiaca and Phlorius and Platzia Flora: similarity 0.50

Leucippe and Clitophon and Phlorius and Platzia Flora: similarity 0.49

Aithiopica and Phlorius and Platzia Flora: similarity 0.49

Callimachus and Chrysorrhoe and Hysmine and Hysminias: similarity 0.49

Leucippe and Clitophon and Callimachus and Chrysorrhoe: similarity 0.48

Callirhoe and Imperios and Margarona: similarity 0.48

Callirhoe and Drosilla and Charicles: similarity 0.48

Leucippe and Clitophon and Aristandros and Callithea: similarity 0.48

Rhodanthe and Dosicles and Aristandros and Callithea: similarity 0.48

Callimachus and Chrysorrhoe and Velthandros and Chrysantza: similarity 0.47

Ephesiaca and Daphnis and Chloe: similarity 0.46

Livistros and Rhodamni and Aristandros and Callithea: similarity 0.46

Callirhoe and Callimachus and Chrysorrhoe: similarity 0.45

Callirhoe and Alexander and Semiramis: similarity 0.45

Callirhoe and Hysmine and Hysminias: similarity 0.45

Imperios and Margarona and Velthandros and Chrysantza: similarity 0.44

Leucippe and Clitophon and Velthandros and Chrysantza: similarity 0.44

Ephesiaca and Callimachus and Chrysorrhoe: similarity 0.43

Callimachus and Chrysorrhoe and Aristandros and Callithea: similarity 0.43

Callirhoe and Aristandros and Callithea: similarity 0.43

Aithiopica and Imperios and Margarona: similarity 0.42

Callirhoe and Phlorius and Platzia Flora: similarity 0.41

Aithiopica and Alexander and Semiramis: similarity 0.41

Ephesiaca and Velthandros and Chrysantza: similarity 0.41

Drosilla and Charicles and Velthandros and Chrysantza: similarity 0.40

Callimachus and Chrysorrhoe and Daphnis and Chloe: similarity 0.40

Rhodanthe and Dosicles and Velthandros and Chrysantza: similarity 0.39

Callirhoe and Daphnis and Chloe: similarity 0.39

Rhodanthe and Dosicles and Alexander and Semiramis: similarity 0.39

Velthandros and Chrysantza and Hysmine and Hysminias: similarity 0.39

Daphnis and Chloe and Phlorius and Platzia Flora: similarity 0.38

Hysmine and Hysminias and Aristandros and Callithea: similarity 0.38

Ephesiaca and Imperios and Margarona: similarity 0.38

Velthandros and Chrysantza and Phlorius and Platzia Flora: similarity 0.36

Alexander and Semiramis and Livistros and Rhodamni: similarity 0.35

Leucippe and Clitophon and Alexander and Semiramis: similarity 0.35

Callimachus and Chrysorrhoe and Imperios and Margarona: similarity 0.35

Leucippe and Clitophon and Imperios and Margarona: similarity 0.34

Alexander and Semiramis and Drosilla and Charicles: similarity 0.34

Alexander and Semiramis and Imperios and Margarona: similarity 0.33

Rhodanthe and Dosicles and Imperios and Margarona: similarity 0.33

Alexander and Semiramis and Hysmine and Hysminias: similarity 0.33

Alexander and Semiramis and Phlorius and Platzia Flora: similarity 0.33

Callimachus and Chrysorrhoe and Alexander and Semiramis: similarity 0.33

Alexander and Semiramis and Velthandros and Chrysantza: similarity 0.32

Imperios and Margarona and Phlorius and Platzia Flora: similarity 0.32

Ephesiaca and Aristandros and Callithea: similarity 0.30

Daphnis and Chloe and Velthandros and Chrysantza: similarity 0.30

Imperios and Margarona and Livistros and Rhodamni: similarity 0.30

Velthandros and Chrysantza and Aristandros and Callithea: similarity 0.29

Ephesiaca and Alexander and Semiramis: similarity 0.28

Daphnis and Chloe and Aristandros and Callithea: similarity 0.28

Alexander and Semiramis and Aristandros and Callithea: similarity 0.27

Drosilla and Charicles and Imperios and Margarona: similarity 0.26

Alexander and Semiramis and Daphnis and Chloe: similarity 0.26

Imperios and Margarona and Hysmine and Hysminias: similarity 0.26

Imperios and Margarona and Daphnis and Chloe: similarity 0.22

Phlorius and Platzia Flora and Aristandros and Callithea: similarity 0.21

Imperios and Margarona and Aristandros and Callithea: similarity 0.13

# APPENDIX D:

Fluctuation Scores for Most Fluctuating Motifs:

Motif 1: Std. Dev. = 0.0165

Motif 0: Std. Dev. = 0.0124

Motif 33: Std. Dev. = 0.0124

Motif 9: Std. Dev. = 0.0118

Motif 11: Std. Dev. = 0.0098

Persistence Scores for Most Persistent Motifs:

Motif 2: Mean = 0.0222

Motif 3: Mean = 0.0169

Motif 5: Mean = 0.0138

Motif 7: Mean = 0.0124

Motif 6: Mean = 0.0118